\documentclass[journal,onecolumn,onehalfspacing]{IEEEtran}

\usepackage[utf8]{inputenc}
\usepackage[T1]{fontenc}
\usepackage[english]{babel}
\usepackage{amsmath, amsfonts, amssymb}
\usepackage{algpseudocode}
\usepackage{algorithm}
\usepackage{array}
\usepackage{textcomp}
\usepackage{stfloats}
\usepackage{verbatim}
\usepackage{graphicx}
\usepackage{cite}
\usepackage{import}
\usepackage{bm}
\usepackage{setspace}
\usepackage{dsfont}
\usepackage{indentfirst}
\usepackage[colorlinks=true, allcolors=blue]{hyperref}

\usepackage{mathtools}
\usepackage{extarrows}
\usepackage{enumitem}
\usepackage{amsthm}
\usepackage{dsfont}
\usepackage[caption=false]{subfig}
\usepackage[export]{adjustbox}
\usepackage{enumitem}
\usepackage{tcolorbox}
\usepackage[all]{xy}
\usepackage{booktabs}
\usepackage{float}
\usepackage{multirow}

\usepackage{tikz}
\usepackage{pgfplots}
\pgfplotsset{compat=newest}

\newcommand{\rbrk}[1]{\left( #1 \right)}
\newcommand{\cbrk}[1]{\left\{ #1 \right\}}

\newcommand{\olsi}[1]{\,\overline{\!{#1}}}

\newcommand{\chartingfunction}{\zeta}
\newcommand{\predictorfunction}{\psi}
\newcommand{\chartdimension}{d}
\newcommand{\channel}{\mathbf{h}}
\newcommand{\velocity}{\mathbf{v}}
\newcommand{\chartembedding}{\mathbf{z}}
\newcommand{\userlocation}{\mathbf{p}}
\newcommand{\numbasestations}{B}
\newcommand{\numantennas}{M}
\newcommand{\numsubcarriers}{W}
\newcommand{\dataset}{\mathcal{D}}
\newcommand{\datasetsize}{D}
\newcommand{\predictionhorizon}{H}
\newcommand{\encoderparameters}{\theta}
\newcommand{\predictorparameters}{\phi}
\newcommand{\emadecay}{\tau}

\usepackage{soul, xcolor}

\usepackage[acronym]{glossaries}

\begin{document}

\newacronym{ml}{ML}{machine learning}
\newacronym{ai}{AI}{artificial intelligence}
\newacronym{csi}{CSI}{channel state information}
\newacronym{jepa}{JEPA}{joint-embedding predictive architecture}
\newacronym{jea}{JEA}{joint-embedding architecture}
\newacronym{ofdm}{OFDM}{orthogonal frequency division multiplex}
\newacronym{pdr}{PDR}{pedestrian dead reckoning}
\newacronym{rnn}{RNN}{recurrent neural network}
\newacronym{ema}{EMA}{exponential moving average}
\newacronym{mlp}{MLP}{multi-layer preceptron}
\newacronym{gru}{GRU}{gated recurrent unit}
\newacronym{lstm}{LSTM}{long short-term memory}
\newacronym{1nn}{$1$--NN}{nearest neighbor}
\newacronym{adp}{ADP}{angle-delay profile}
\newacronym{wjepa}{W-JEPA}{}

\glsdisablehyper

\title{Learning Latent Wireless Dynamics from \\ Channel State Information
}

\author{
Charbel Bou Chaaya, Abanoub M. Girgis and Mehdi Bennis
\thanks{%
The authors are with the Centre for Wireless Communications, University of Oulu, Finland (email: charbel.bouchaaya@oulu.fi; abanoub.pipaoy@oulu.fi; mehdi.bennis@oulu.fi).
}
}

\maketitle

\begin{abstract}
In this work, we propose a novel data-driven \gls{ml} technique to model and predict the dynamics of the wireless propagation environment in latent space.
Leveraging the idea of channel charting, which learns compressed representations of high-dimensional \gls{csi}, we incorporate a predictive component to capture the dynamics of the wireless system.
%
%
Hence, we jointly learn a channel encoder that maps the estimated \gls{csi} to an appropriate latent space, and a predictor that models the relationships between such representations.
Accordingly, our problem boils down to training a \gls{jepa} that simulates the latent dynamics of a wireless network from \gls{csi}.
%
%
We present numerical evaluations on measured data and show that the proposed \gls{jepa} displays a two-fold increase in accuracy over benchmarks, for longer look-ahead prediction tasks.
\end{abstract}


\glsresetall

\section{Introduction}
\IEEEPARstart{T}{he} role of \gls{ai} in the design and optimization of next-generation wireless networks is of paramount importance.
The predictions capabilities of \gls{ai} will allow the optimization of protocols at various communication layers.
At the physical layer, \gls{ai} will help address challenging radio resource management tasks, stemming from the increasing number of antennas and complex nature of the environment to ensure seamless user experiences~\cite{park2022extreme}.

A key technique that allows learning agents to generalize in such complex environments is rooted in the so-called `world model'~\cite{lecun2022path}.
As the name suggests, a `world model' is an accumulation of an agent's knowledge about the environment's dynamics, learned in a self-supervised fashion from the agent's interactions, that helps in predicting the consequences of actions.
Subsequently, such model can be used to forecast future states of the environment as a function of action sequences.

In this work, we present a new self-supervised learning technique to learn latent representations of the wireless \gls{csi} dynamics.
Particularly, our aim is to predict future channel embeddings from a given channel realization and a suitable conditioning variable, i.e. the user's velocity.
Our study builds on the recent progress in channel charting, a self-supervised learning method that aims to compress the \gls{csi} manifold into a low-dimensional representation space, denoted as channel chart~\cite{studer2018channel}.
While it has shown a promising performance on a multitude of wireless problems~\cite{ferrand2023wireless}, channel charting is not concerned with predicting future latent states, by instead learning a mapping from raw \gls{csi} data to their latent representations.
We seek to augment the channel charting framework with a predictive capability, i.e. a feature that can suppress the need to estimate future channel realizations.
To learn such latent representation of the wireless environment dynamics, we propose a \gls{jepa} for \gls{csi} data, which is a trainable module that simulates the dynamics of the relevant information in the environment~\cite{lecun2022path}.
While various \gls{jepa} models have shown superior results on computer vision problems~\cite{assran2023self}, its performance on wireless modalities has not been investigated.
We present a set of comprehensive numerical results, that showcase the advantages of our approach, reaching more than $40\%$ accuracy increase over baselines.


\section{Preliminaries and Related Works}\label{section:ccjepa_2}

\subsection{Joint-Embedding Predictive Architectures}
\begin{figure*}
    \centering
    \subfloat[Joint-embedding architectures.]
    {\includegraphics[width=.27\textwidth,valign=b]{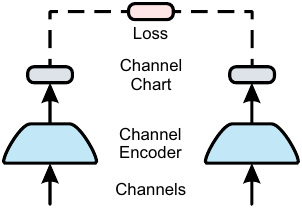}\label{fig:cc_jepa_paper_jea}}
    \hfill
    \subfloat[Generative architectures.]
    {\includegraphics[width=.28\textwidth,valign=b]{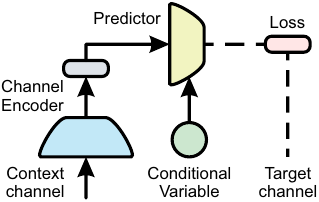}\label{fig:cc_jepa_paper_gen}}
    \hfill
    \subfloat[Joint-embedding predictive architectures.]
    {\includegraphics[width=.3\textwidth,valign=b]{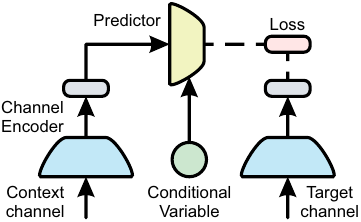}\label{fig:cc_jepa_paper_jepa}}
    \hfill
    \caption{Common architectures for self-supervised learning.}
    \label{fig:cc_jepa_paper_ssl}
\end{figure*}

Prevailing approaches to self-supervised learning can be categorized under two groups: \glspl{jea} and generative architectures.
The objective of \glspl{jea} (Fig.~\ref{fig:cc_jepa_paper_jea}) is to train a model to learn similar embeddings for comparable inputs, and distinct embeddings for contrasting inputs.
A pitfall of \glspl{jea} is representation collapse, where the model maps all inputs to a single output.
For example, to avoid such trivial solutions, contrastive losses explicitly push dissimilar embeddings.
On the other hand, generative methods (Fig.~\ref{fig:cc_jepa_paper_gen}) aim to reconstruct a target signal from a given input signal.
The reconstruction is facilitated by some conditioning information that indicates the relation between them.
Although generative architectures do not suffer from representation collapse, a fundamental issue in their design is the need to fully predict every aspect of the target signal, which is generally inconvenient and/or cumbersome.

\Glspl{jepa} (Fig.~\ref{fig:cc_jepa_paper_jepa}) seek to predict the representation of a target signal from the representation of an input.
Technically, \gls{jepa} infers only the `relevant information' of the target signal, instead of regenerating it.
Similarly to generative architectures, the prediction is done through a predictor network conditioned on additional information.
However, the loss of \gls{jepa} is calculated in the latent space, instead of raw data space.

\glspl{jepa} have been mostly applied to computer vision modalities such as images~\cite{assran2023self}, videos~\cite{bardes2024revisiting}, point clouds~\cite{saito2024point} and audio signals~\cite{fei2023jepa}.
The authors of \cite{girgis2024time} used \gls{jepa} to learn the dynamics of a remote control system whose state is captured by an image, to minimize the communication overhead between a sensor and an actuator.

\subsection{Channel Charting}
Channel charting is a self-supervised learning method whose objective is to learn a map from high dimensional \gls{csi} data to a low dimensional space, referred to as the channel chart.
The chart is a latent representation of the wireless channels that preserves spatial neighborhoods, i.e. channels originating from nearby locations are mapped to adjacent chart points, and vice-versa.
As such, a chart point is understood as a pseudo-location of a user, obtained by processing its \gls{csi}.
Formally, one or many base stations collect a dataset of channels $\cbrk{\mathbf{h}_i}$ and seek a forward charting function~\cite{studer2018channel}:
\begin{equation}
    \begin{matrix}
        \chartingfunction: &\mathbb{C}^{N} \; & \longrightarrow \; &\mathbb{R}^\chartdimension \\
        &\channel \; & \longmapsto \; &\chartembedding
    \end{matrix}
\end{equation}
where $N$ is the channel's dimension, and $\chartdimension$ is a user-defined parameter usually set to $2$.
The map is learned such that, for distinct channels $\channel, \channel^\prime$ estimated from user locations $\userlocation, \userlocation^\prime$, it satisfies:
\begin{equation}
    \userlocation \approx \userlocation^\prime \quad \Longleftrightarrow \quad \chartingfunction\rbrk{\channel} \approx \chartingfunction\rbrk{\channel^\prime}.
\end{equation}

We note that unlike supervised fingerprinting methods that require \gls{csi} data labeled by the user's position, channel charting is a self-supervised technique that processes only estimated wireless channels.
It is also worth mentioning that although the original requirement of channel charting is to conserve neighborhoods locally, recent studies attempted at learning a globally robust channel chart that preserves the overall spatial geometry~\cite{stephan2024angle}, \cite{stahlke2024velocity}.

\section{System Model and Problem Formulation}\label{section:ccjepa_3}
We consider the uplink of a wireless system, where a base station consisting of $\numbasestations$ antenna arrays equipped with $\numantennas$ antennas each, serves a single antenna user.
We first assume that the base station estimates the user's channel coefficients $\channel \in \mathbb{C}^{\numbasestations \times \numantennas \times \numsubcarriers}$ over $\numsubcarriers$ \gls{ofdm} subcarriers.

Generally, given a dataset of estimated \gls{csi}, most common approaches to channel charting proceed by first defining a channel dissimilarity measure, and then training a \gls{jea}, such as triplet or siamese networks, to learn channel embeddings that are separated by the original distance of their corresponding channel inputs.
Then, for a given estimated channel, the base station can readily find its embedding in the chart, which is typically used to solve a downstream radio resource management task.

Our objective in this work diverges from the conventional channel charting approach.
While we aim to learn a low-dimensional \gls{csi} representation space similar to a channel chart, we do not explicitly define a channel distance. 
We jointly seek a channel encoder that embeds wireless channels in a low dimensional space, such that we predict one channel embedding from another given an extra variable that captures the relation between the channels.
In other words, our goal is to map high dimensional channels to a latent space (similar to a channel chart), such that we can replace the expensive estimation of subsequent channels, with a simpler estimation of the conditional variable that guides the prediction in the latent space.
In this context, learning the channel chart is more of a byproduct of our primary goal.
As our channel encoder serves a similar purpose as a channel charting encoder, we can interpret the obtained embedding space as a channel chart of pseudo-locations.
Therefore, a candidate variable that relates subsequent channel representations is the user's velocity.
In fact, our intuition is that consecutive channel measurements, originating from a user movement between consecutive locations with a given velocity, should be mapped to the chart's pseudo-locations such that a subsequent embedding is inferred from a former one given the velocity\footnote{Note that our approach is essentially distinct from~\cite{stahlke2024velocity}, which utilizes the velocity to define a distance measure between channels and learn the charting function.}.
%
%
Thus, we replace the channel estimation task (whenever unnecessary) with the much simpler velocity estimation.
Henceforth, we assume that the base station continuously tracks the user's velocity\footnote{Nowadays, velocity information is integrated in a wide range of devices, and can be easily obtained using odometry or \gls{pdr} algorithms.}.

Our goal is to learn a mapping $\chartingfunction$ that transforms a channel observation $\channel_n$ estimated at time slot $n$ to a low dimensional embedding $\chartembedding_n = \chartingfunction\rbrk{\channel_n}$, from which we can infer the representation of a subsequent channel at slot $n+j$ given a sequence of the user's velocity $\rbrk{\velocity_n, \dots, \velocity_{n+j-1}}$ through another mapping $\predictorfunction$, i.e. $\chartembedding_{n+j} = \predictorfunction\rbrk{\chartembedding_n \mid \velocity_n, \dots, \velocity_{n+j-1}}$.
We recapitulate this task in the following schematic, where $\predictionhorizon$ is the prediction horizon:
\begin{equation*}
    \xymatrix{ \channel_{n} \ar@{-->}[rrrr]^-{\rbrk{\velocity_{n}, \dots, \velocity_{n+\predictionhorizon-1}}} \ar[d]^{\chartingfunction} && \ar@{}[d]^(.1){}="a"^(.95){}="b" \ar "a";"b" && \rbrk{\channel_{n+1}, \dots, \channel_{n+\predictionhorizon}} \ar@{-->}[d]  \\ \chartembedding_{n} \ar[rrrr]|-{\predictorfunction} &&&& \rbrk{\chartembedding_{n+1}, \dots, \chartembedding_{n+\predictionhorizon}} }
\end{equation*}

To proceed, we consider that the base station collects a data set $\dataset = \cbrk{\channel_i, \velocity_i}_{i=1}^{\datasetsize}$ consisting of $\datasetsize$ consecutive channel samples with the corresponding user velocity.

\section{Proposed Solution}\label{section:ccjepa_4}
We solve the aforementioned problem using a suitable \gls{jepa}, whose structure is described in the following.

\begin{figure}
    \centering
    \includegraphics[width=.475\textwidth]{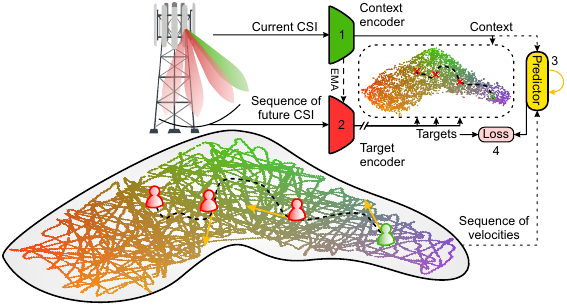}
    \caption{System model and our proposed method.}
    \label{fig:cc_jepa_paper_system}
\end{figure}

\subsection{Wireless JEPA}
Our \gls{jepa}, dubbed \gls{wjepa}, comprises the following components, as depicted in Fig.~\ref{fig:cc_jepa_paper_system}.
\subsubsection{Context}
For an arbitrary time slot $n$, we feed the estimated channel $\channel_n$ to the encoder $\chartingfunction_{\encoderparameters}$, parametrized by a set $\encoderparameters$ of learnable parameters. We note that the encoder mimics a channel chart function whose output is $2$-dimensional vector, $\chartembedding_n = \chartingfunction_{\encoderparameters}\rbrk{\channel_n}$, representing the user's pseudo-location.
\subsubsection{Targets}
Given a prediction horizon $\predictionhorizon$, we feed the sequence of consecutive channels $\rbrk{\channel_{n+1}, \dots, \channel_{n+\predictionhorizon}}$ to the target encoder $\chartingfunction_{\olsi{\encoderparameters}}$, whose output is a sequence of pseudo-locations $\rbrk{\chartembedding_{n+1}, \dots, \chartembedding_{n+\predictionhorizon}}$.
\subsubsection{Prediction}
The goal of \gls{jepa} is to predict the target's representation from the context's representation.
Hence, our prediction problem reduces to a sequence forecasting problem, i.e. an autoregressive prediction of consecutive chart points given the user's real velocity.
Hence, we parametrize the predictor as a \gls{rnn} with parameters $\predictorparameters$, as they are known to perform well on time series prediction.
The predictor takes as input the context encoder's output $\chartembedding_n$ and a sequence of raw velocities ${\rbrk{\velocity_{n}, \dots, \velocity_{n+\predictionhorizon-1}}}$.
It then outputs a prediction sequence $\predictorfunction_{\predictorparameters}\rbrk{\chartembedding_n \mid \velocity_n, \dots, \velocity_{n+\predictionhorizon-1}} = \rbrk{\hat{\chartembedding}_{n+1}, \dots, \hat{\chartembedding}_{n+\predictionhorizon}}$.
\subsubsection{Loss}
Our training loss is the average $\ell_2$ distance between the predicted channel embeddings, and the target embeddings.

In a nutshell, the encoder's and predictor's parameters $\encoderparameters$ and $\predictorparameters$ are jointly learned through gradient-based optimization, as follows:
\begin{equation}\label{eq:ccjepa_paper_problem}
    \underset{\encoderparameters, \, \predictorparameters}{\text{minimize}} \qquad \frac{1}{\datasetsize\predictionhorizon} \sum_{n=1}^{\datasetsize} \, \sum_{t=1}^{\predictionhorizon} \, \left\lVert 
    \hat{\chartembedding}_{n+t} - \chartembedding_{n+t}
    \right\rVert^2_2,
\end{equation}
while the target encoder's parameters are updated, at each training step, by an \gls{ema} of the context encoder's parameters with a decay rate $\emadecay$:
\begin{equation}\label{eq:ccjepa_paper_ema}
    \olsi{\encoderparameters} \,\leftarrow\, \emadecay \olsi{\encoderparameters} + \rbrk{1-\emadecay} \encoderparameters.
\end{equation}
More precisely, the target encoder has the same network architecture as the online context encoder, and they both share the same parameters at the start of training.
As the target encoder provides the target `labels' to train the context encoder, we stop the gradient flow through its branch to prevent representation collapse, and update its weights using \eqref{eq:ccjepa_paper_ema}.
The idea is to train the encoder to produce high quality representations from which the predictor infers the targets.
Then, instead of directly updating the target's parameters by the context's weights, we use a slow \gls{ema} update, and iterate this procedure to improve the representation quality of the encoder~\cite{grill2020bootstrap}.

Once trained, the proposed network can be utilized to map a given channel to its chart embedding using the learned encoder, and use the predictor to autoregressively infer the subsequent sequence of channel representations, by conditioning on the sequence of estimated velocities.

\subsection{Two-stage Curriculum Learning}
In general, it is possible to train our proposed \gls{wjepa} network from scratch following \eqref{eq:ccjepa_paper_problem}.
However, as both representation learning and prediction tasks are challenging, jointly learning them leads to sub-optimal results.
Inspired by the \gls{ml} literature on curriculum learning~\cite{bengio2009curriculum}, we propose a simple two-stage training method.
Curriculum learning is a \gls{ml} approach where the model learns tasks of increasing difficulty during its training process, gradually modulating the simple principles learned on easier tasks to more refined and complex concepts.
In this context, directly exposing \gls{jepa} to learn the joint channel embedding and prediction tasks is a difficult problem that can be cast as two sub-problems.

In the first step, the encoder is pre-trained in a typical channel charting scheme on a small subset of estimated channels in $\dataset$.
Any channel distance from the literature and learning pattern could be used, such as contrastive learning.
By doing so, we induce a good inductive bias that simplifies the ensuing joint encoding-prediction task, as the encoder learns clustered channel representations, from which the prediction task becomes much easier.
In the second stage, the encoder is then trained jointly with the predictor to minimize the \gls{jepa} loss.
We numerically show the impact of this approach in the following section.
We emphasize that the optional pre-training step is also self-supervised as we only make use of unlabeled \gls{csi} data.

\section{Numerical Results}\label{section:ccjepa_5}

\begin{figure*}
    \centering
    \subfloat[Ground truth locations]{\parbox[t][4cm][t]{0.26\textwidth}{
        \begin{tikzpicture}[baseline,remember picture]
    \begin{axis}[
        width=0.18\textwidth,
        height=0.18\textwidth,
        scale only axis,
        xmin=-12.5,
        xmax=2.5,
        ymin=-14.5,
        ymax=-1.5,
        xlabel = {$x$ [m]},
        ylabel = {$y$ [m]},
        tick label style={font=\scriptsize},
        major tick length=2pt,
        every tick/.style={thin},
        label style={font=\footnotesize},
        ylabel shift = -8 pt,
        xlabel shift = -4 pt,
        xtick={-10, -6, -2, 2},
        ytick={-14, -10, -6, -2},
    ]
        \addplot[thick,blue] graphics[xmin=-12.5,xmax=2,ymin=-14.5,ymax=-1.5] {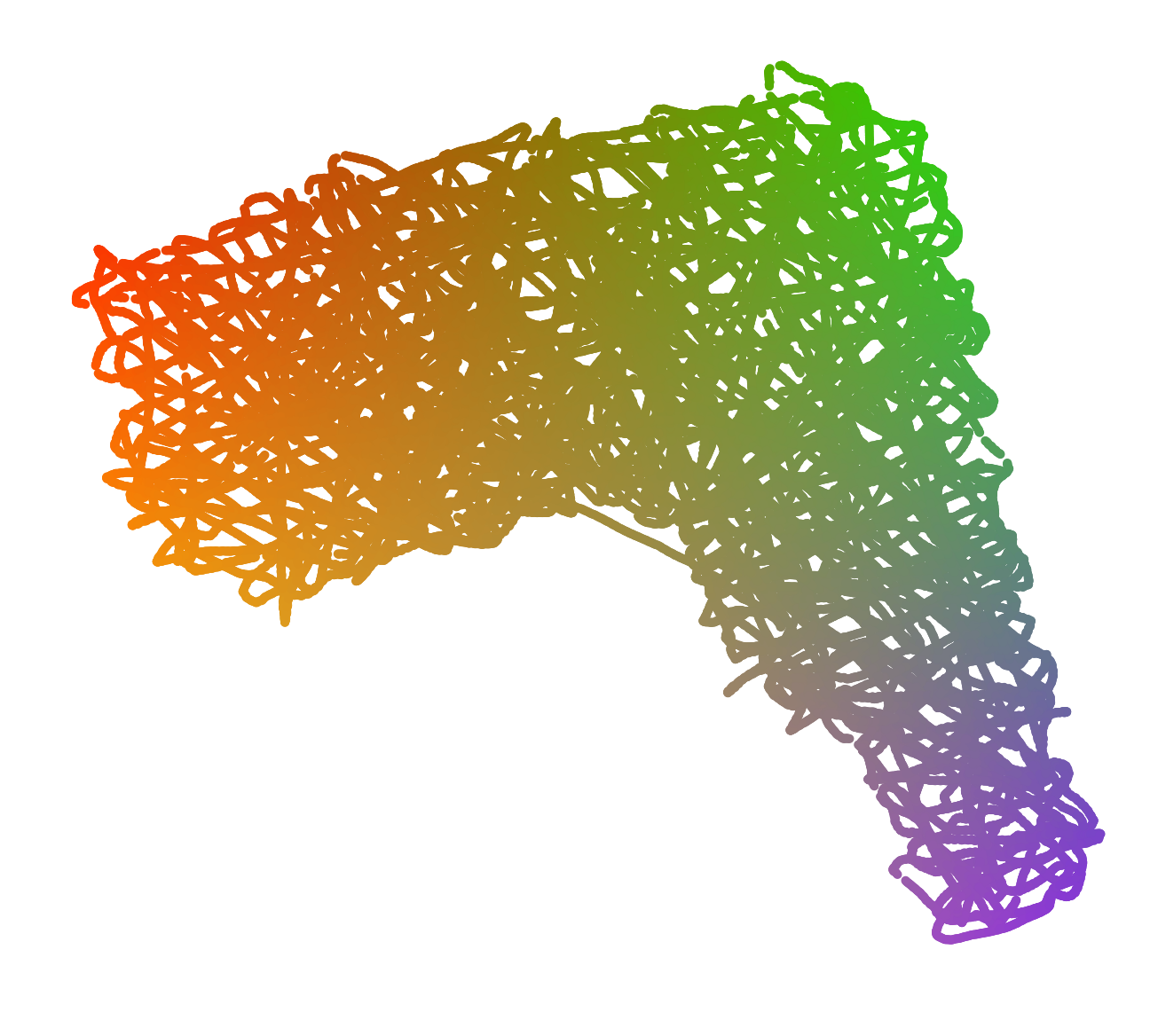};
    \end{axis}
\end{tikzpicture}
        \label{fig:cc_jepa_paper_gt_locations}}}
    \hfill
    \subfloat[No pre-training]{\parbox[t][4cm][t]{0.22\textwidth}{
        \begin{tikzpicture}[baseline,remember picture]
    \begin{axis}[
        width=0.18\textwidth,
        height=0.18\textwidth,
        scale only axis,
        xmin=-1.5,
        xmax=1.5,
        ymin=-1,
        ymax=1,
        xlabel = {\phantom{$x$ [m]}},
        ylabel = {\phantom{$y$ [m]}},
        tick label style={font=\scriptsize},
        label style={font=\small},
        ylabel shift = -8 pt,
        xlabel shift = -4 pt,
        ticks=none,
    ]
        \addplot[thick,blue] graphics[xmin=-1.5,xmax=1.5,ymin=-1,ymax=1] {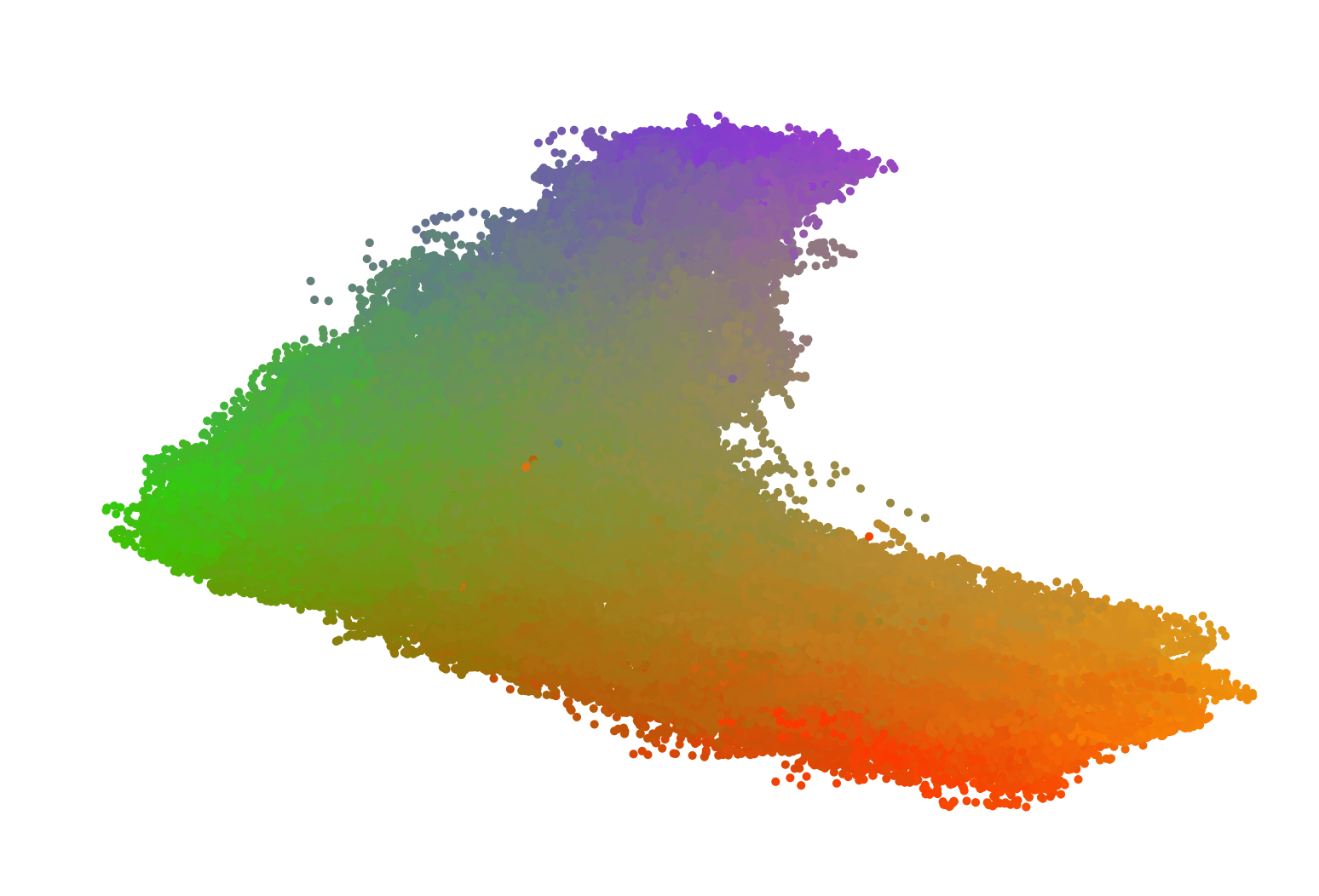};
    \end{axis}
\end{tikzpicture}
        \label{fig:cc_jepa_paper_random_chart}}}
    \hfill
    \subfloat[Pre-trained with $d_{\text{ADP}}$]{\parbox[t][4cm][t]{0.22\textwidth}{
        \begin{tikzpicture}[baseline,remember picture]
    \begin{axis}[
        width=0.18\textwidth,
        height=0.18\textwidth,
        scale only axis,
        xmin=-25,
        xmax=24,
        ymin=-18,
        ymax=25,
        xlabel = {\phantom{$x$ [m]}},
        ylabel = {\phantom{$y$ [m]}},
        tick label style={font=\small},
        label style={font=\small},
        ylabel shift = -8 pt,
        xlabel shift = -4 pt,
        ticks=none,
    ]
        \addplot[thick,blue] graphics[xmin=-25,xmax=24,ymin=-18,ymax=25] {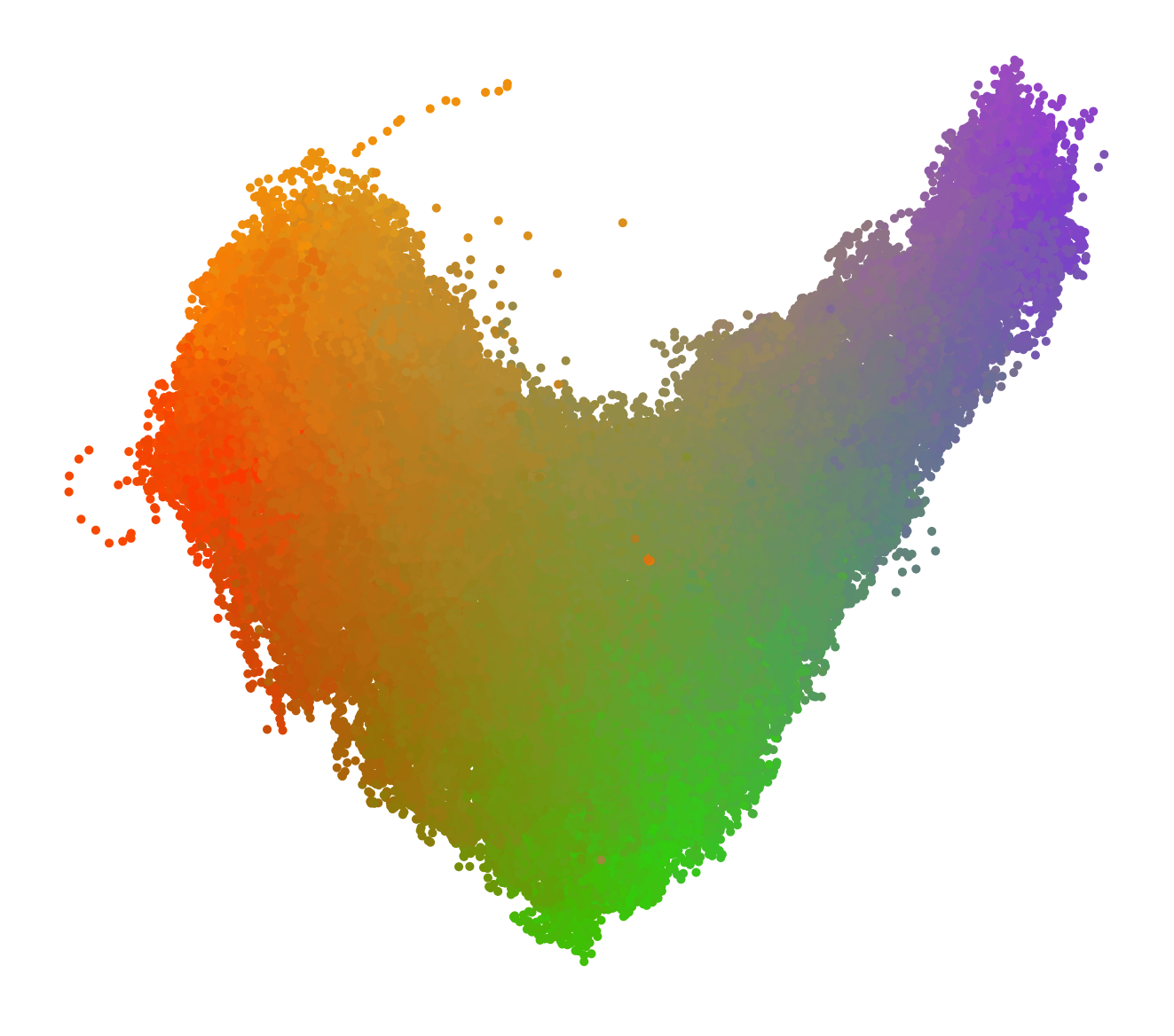};
    \end{axis}
\end{tikzpicture}
        \label{fig:cc_jepa_paper_local_chart}}}
    \hfill
    \subfloat[Pre-trained with $d_{\text{G-fuse}}$]{\parbox[t][4cm][t]{0.22\textwidth}{
        \begin{tikzpicture}[baseline,remember picture]
    \begin{axis}[
        width=0.18\textwidth,
        height=0.18\textwidth,
        scale only axis,
        xmin=-450,
        xmax=250,
        ymin=-250,
        ymax=300,
        xlabel = {\phantom{$x$ [m]}},
        ylabel = {\phantom{$y$ [m]}},
        tick label style={font=\small},
        label style={font=\small},
        ylabel shift = -8 pt,
        xlabel shift = -4 pt,
        ticks=none,
    ]
        \addplot[thick,blue] graphics[xmin=-450,xmax=250,ymin=-250,ymax=300] {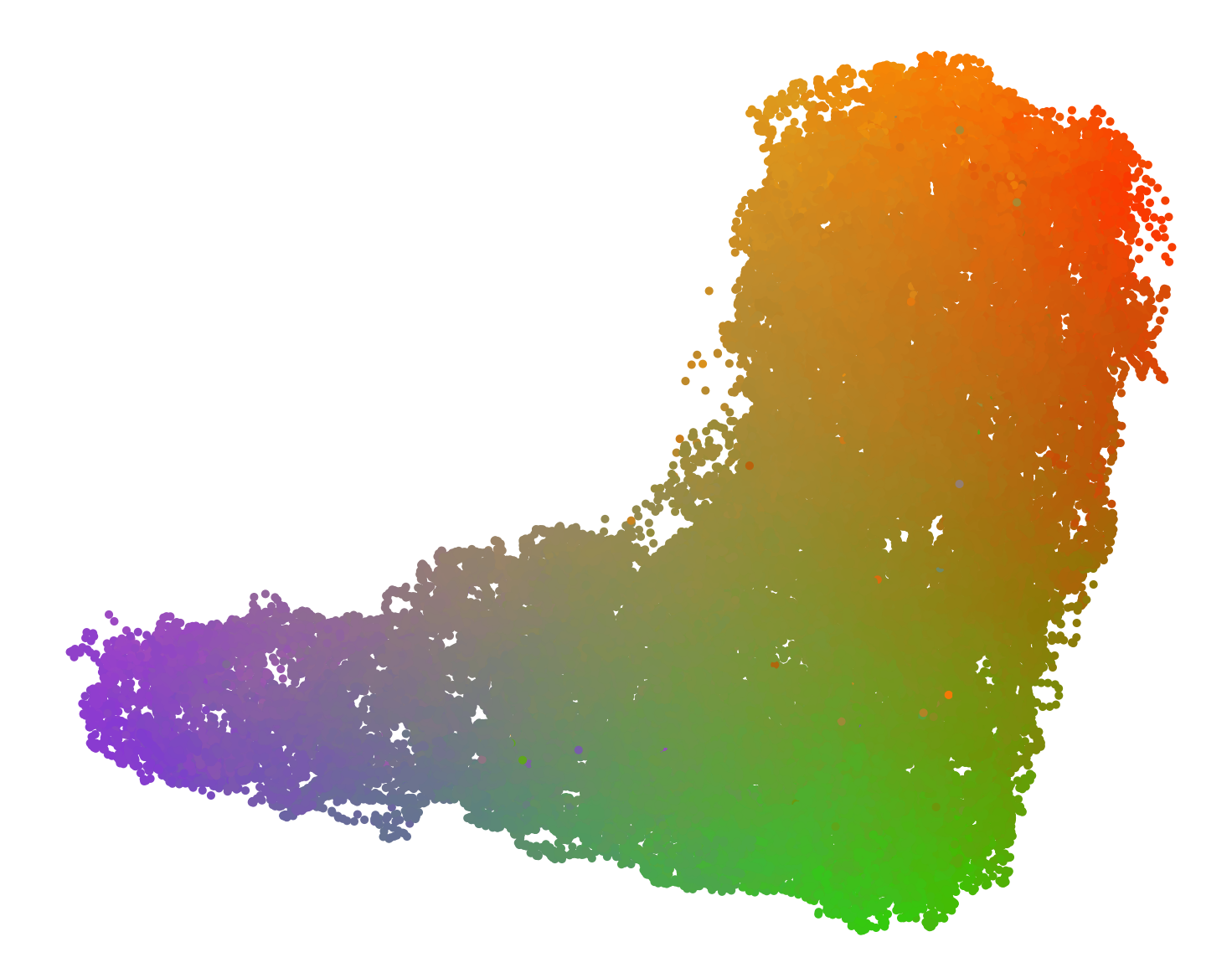};
    \end{axis}
\end{tikzpicture}
        \label{fig:cc_jepa_paper_global_chart}}}
    \caption{Impact of pre-training the encoder on the learned channel charts.}
    \label{fig:cc_jepa_paper_charts}
\end{figure*}

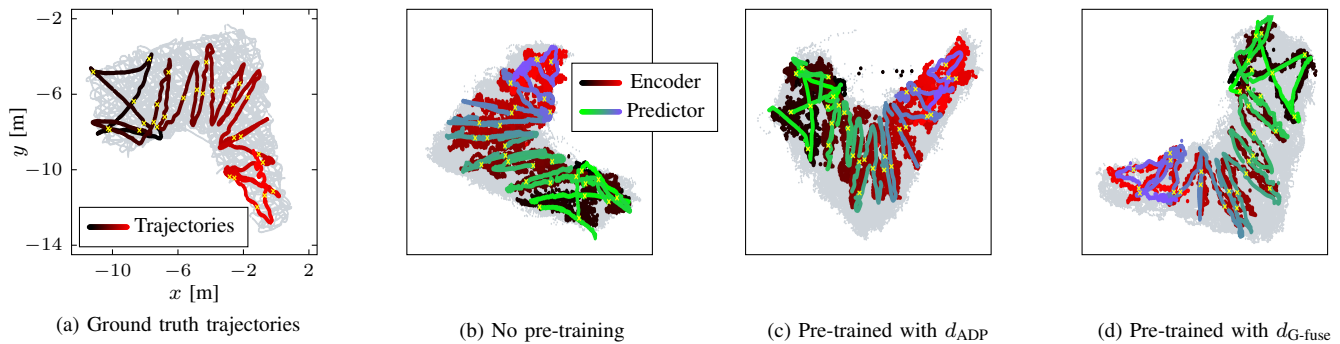
\begin{figure*}
    \centering
    \subfloat[Ground truth trajectories]{\parbox[t][4cm][t]{0.26\textwidth}{
        \begin{tikzpicture}[baseline,remember picture]
    \begin{axis}[
        width=0.18\textwidth,
        height=0.18\textwidth,
        scale only axis,
        xmin=-12.5,
        xmax=2.5,
        ymin=-14.5,
        ymax=-1.5,
        xlabel = {$x$ [m]},
        ylabel = {$y$ [m]},
        tick label style={font=\scriptsize},
        major tick length=2pt,
        every tick/.style={thin},
        label style={font=\footnotesize},
        ylabel shift = -8 pt,
        xlabel shift = -4 pt,
        xtick={-10, -6, -2, 2},
        ytick={-14, -10, -6, -2},
        legend style={
          fill opacity=0.5,
          draw opacity=1,
          text opacity=1,
          at={(0.03,0.03)},
          anchor=south west,
          draw=black,
          font=\footnotesize
        },
        trajectory/.style={
          legend image code/.code={
            \foreach \x in {0,1,2,...,100}
                \draw[black!\x!red,only marks,mark=*,mark size=0.75,mark options={fill opacity=1}] plot coordinates {(1mm-\x/7,0cm)};
          }
        }
    ]
        \addlegendimage{trajectory}
        \legend{Trajectories}
        \addplot[thick,blue] graphics[xmin=-12.5,xmax=2,ymin=-14.5,ymax=-1.5] {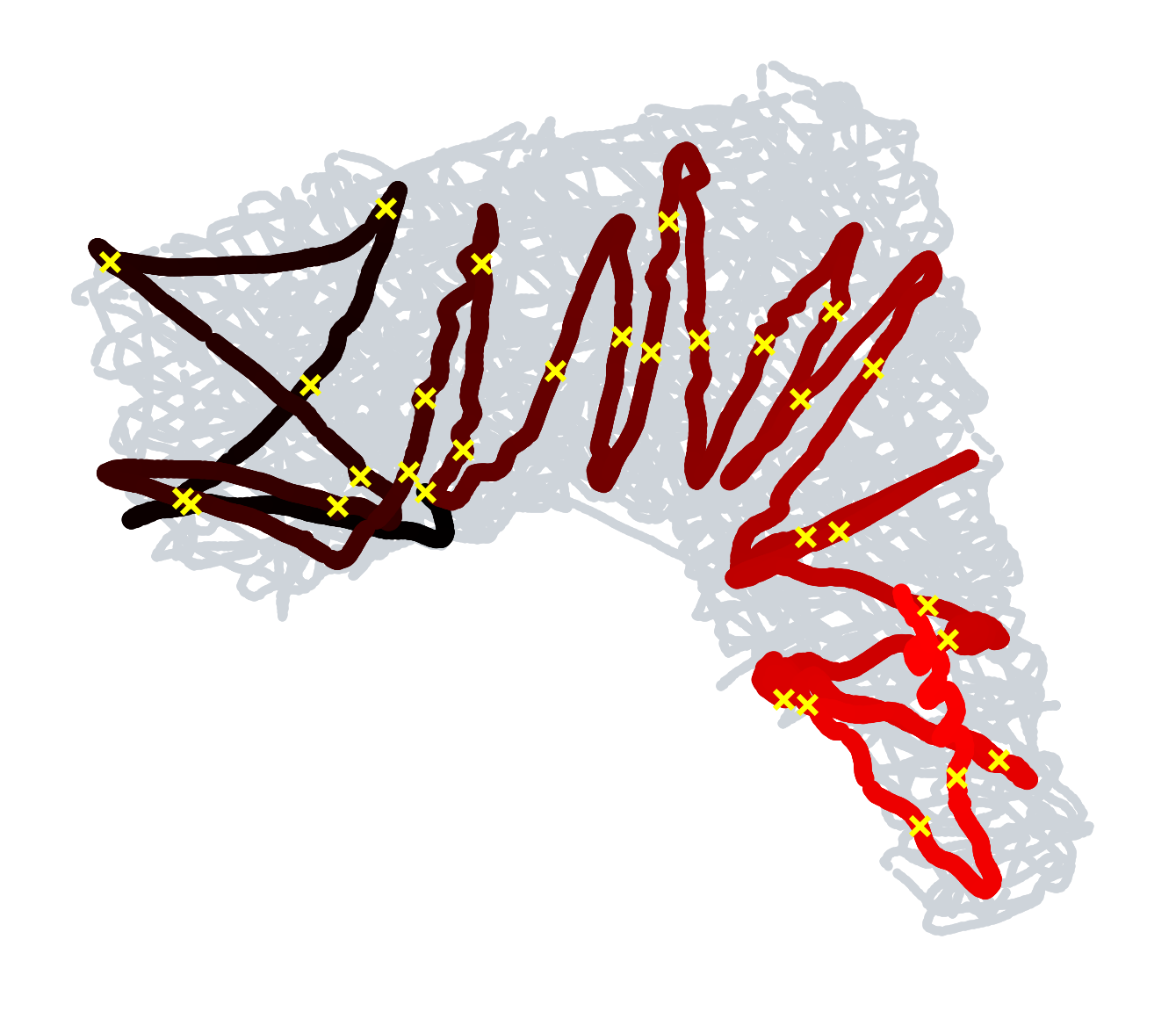};
    \end{axis}
\end{tikzpicture}
        \label{fig:cc_jepa_paper_gt_traj}}}
    \hfill
    \subfloat[No pre-training]{\parbox[t][4cm][t]{0.22\textwidth}{
        \begin{tikzpicture}[baseline,remember picture]
    \definecolor{pregreen}{RGB}{0,255,0}
    \definecolor{preblue}{RGB}{128,82,255}
    \begin{axis}[
        width=0.18\textwidth,
        height=0.18\textwidth,
        scale only axis,
        xmin=-1.5,
        xmax=1.5,
        ymin=-1,
        ymax=1,
        xlabel = {\phantom{$x$ [m]}},
        ylabel = {\phantom{$y$ [m]}},
        tick label style={font=\scriptsize},
        label style={font=\footnotesize},
        ylabel shift = -8 pt,
        xlabel shift = -4 pt,
        ticks=none,
        legend style={
          fill opacity=1,
          draw opacity=1,
          text opacity=1,
          at={(0.67,0.5)},
          anchor=south west,
          draw=black,
          font=\footnotesize
        },
        encoder/.style={
          legend image code/.code={
            \foreach \x in {0,1,2,...,100}
                \draw[black!\x!red,only marks,mark=*,,mark size=0.75,mark options={fill opacity=1}] plot coordinates {(1mm-\x/7,0cm)};
          }
        },
        predictor/.style={
          legend image code/.code={
            \foreach \x in {0,1,2,...,100}
                \draw[pregreen!\x!preblue,only marks,mark=*,,mark size=0.75,mark options={fill opacity=1}] plot coordinates {(1mm-\x/7,0cm)};
          }
        }
    ]
        \addlegendimage{encoder}
        \addlegendimage{predictor}
        \legend{Encoder, Predictor}
        \addplot[thick,blue] graphics[xmin=-1.5,xmax=1.5,ymin=-1,ymax=1] {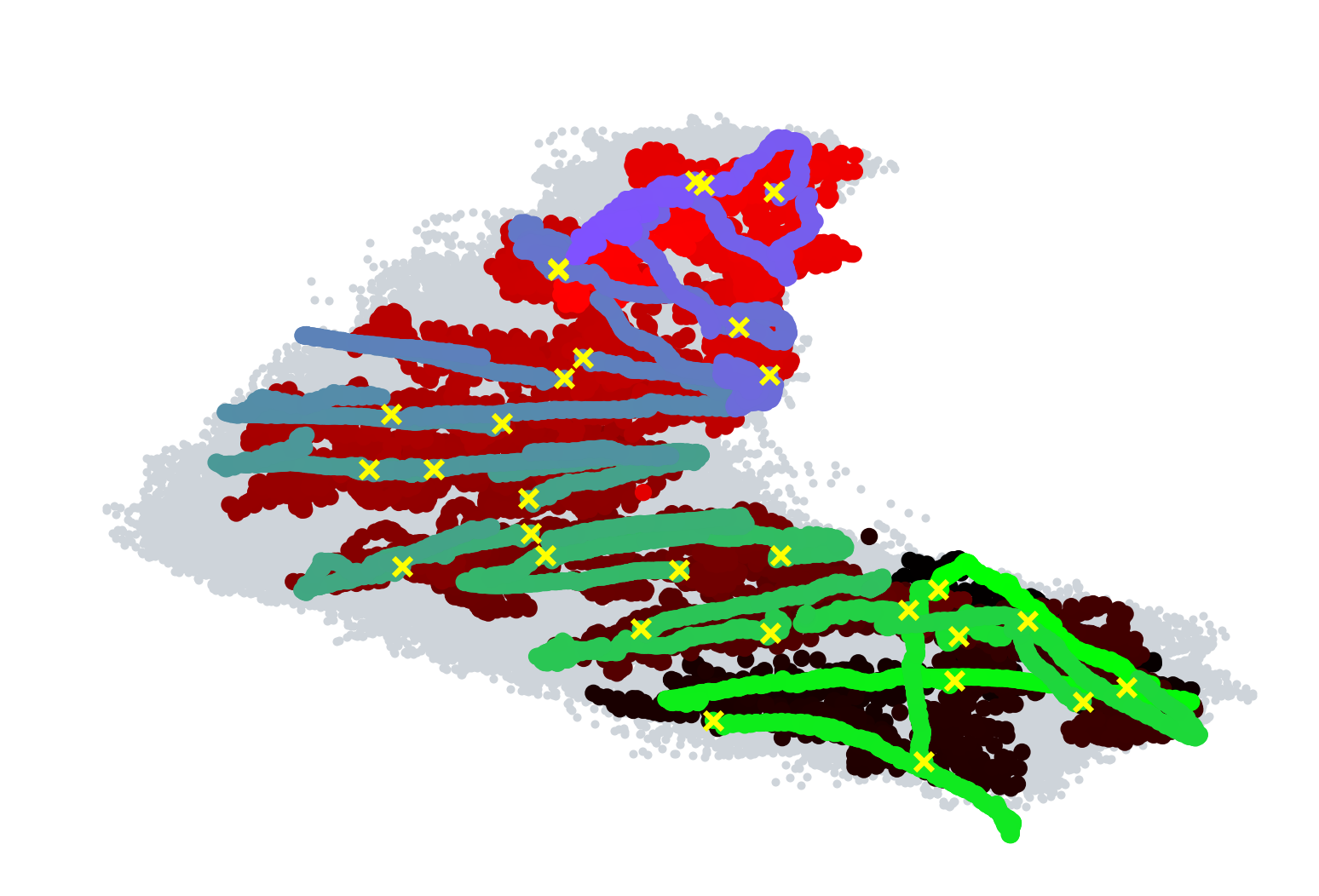};
    \end{axis}
\end{tikzpicture}
        \label{fig:cc_jepa_paper_random_traj}}}
    \hfill
    \subfloat[Pre-trained with $d_{\text{ADP}}$]{\parbox[t][4cm][t]{0.22\textwidth}{
        \begin{tikzpicture}[baseline,remember picture]
    \begin{axis}[
        width=0.18\textwidth,
        height=0.18\textwidth,
        scale only axis,
        xmin=-25,
        xmax=24,
        ymin=-18,
        ymax=25,
        xlabel = {\phantom{$x$ [m]}},
        ylabel = {\phantom{$y$ [m]}},
        tick label style={font=\small},
        label style={font=\small},
        ylabel shift = -8 pt,
        xlabel shift = -4 pt,
        ticks=none,
    ]
        \addplot[thick,blue] graphics[xmin=-25,xmax=24,ymin=-18,ymax=25] {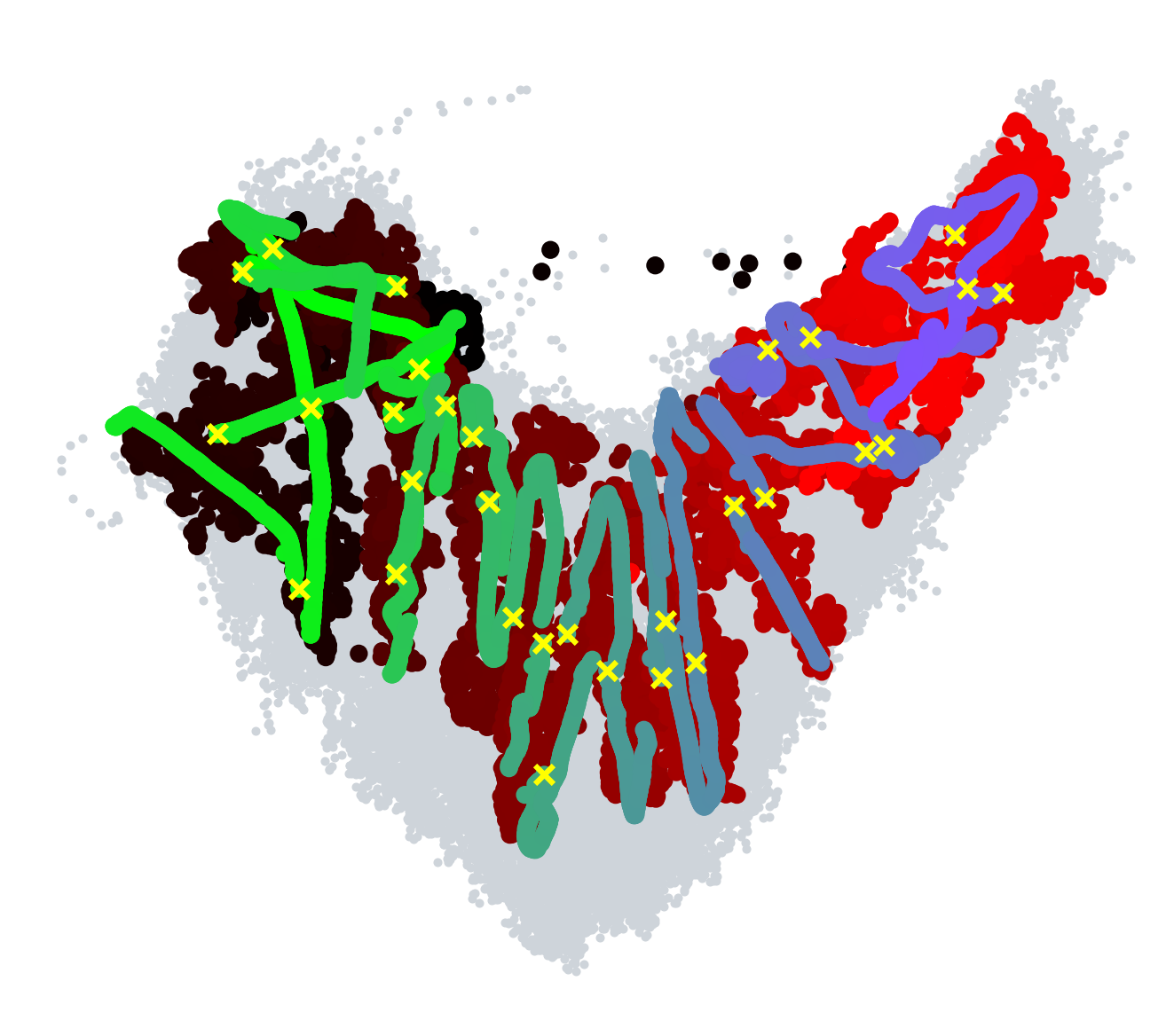};
    \end{axis}
\end{tikzpicture}
        \label{fig:cc_jepa_paper_local_traj}}}
    \hfill
    \subfloat[Pre-trained with $d_{\text{G-fuse}}$]{\parbox[t][4cm][t]{0.22\textwidth}{
        \begin{tikzpicture}[baseline,remember picture]
    \begin{axis}[
        width=0.18\textwidth,
        height=0.18\textwidth,
        scale only axis,
        xmin=-450,
        xmax=250,
        ymin=-250,
        ymax=300,
        xlabel = {\phantom{$x$ [m]}},
        ylabel = {\phantom{$y$ [m]}},
        tick label style={font=\small},
        label style={font=\small},
        ylabel shift = -8 pt,
        xlabel shift = -4 pt,
        ticks=none,
    ]
        \addplot[thick,blue] graphics[xmin=-450,xmax=250,ymin=-250,ymax=300] {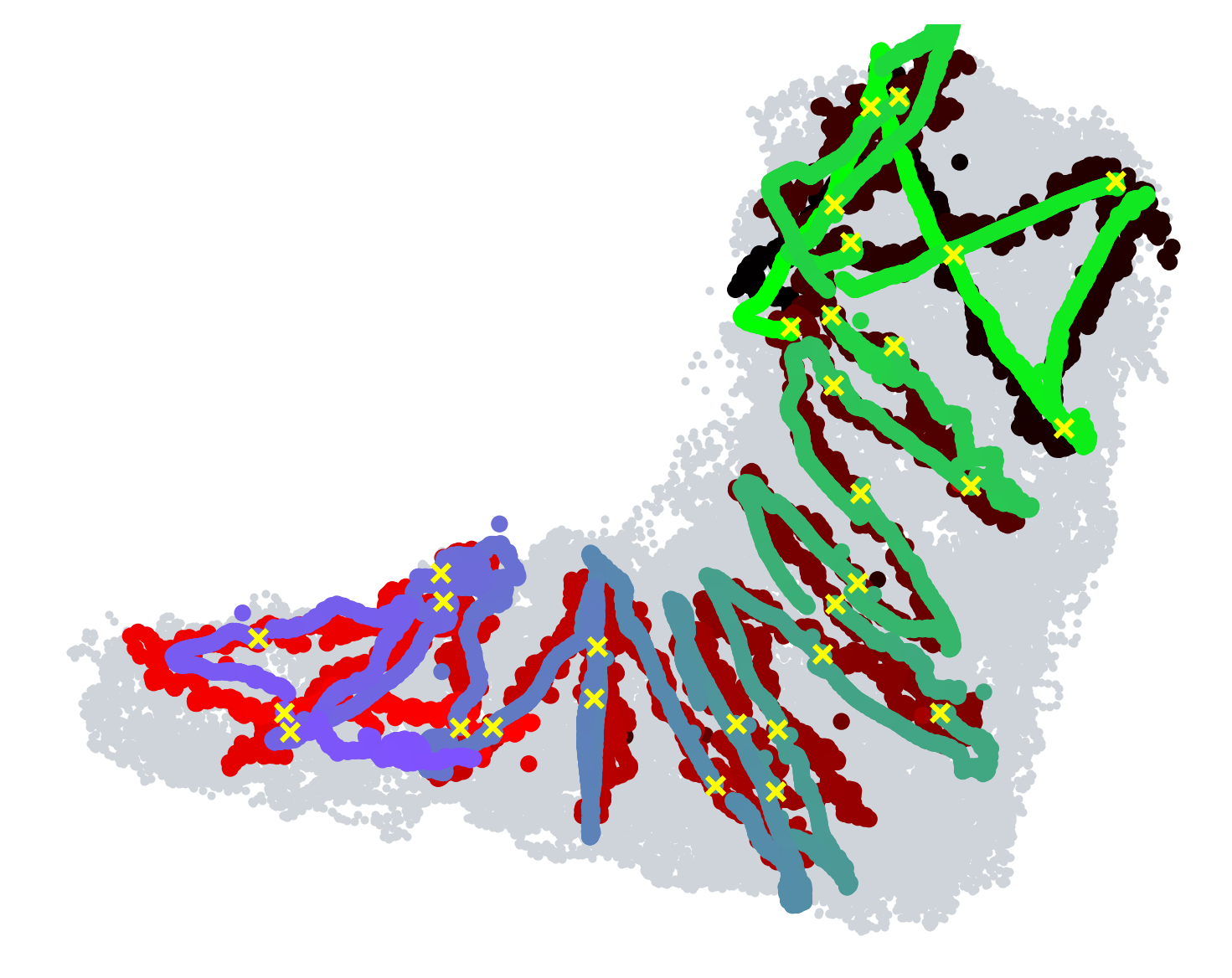};
    \end{axis}
\end{tikzpicture}
        \label{fig:cc_jepa_paper_global_traj}}}
    \caption{The user's dynamics as modelled by the predictor in the channel charts.}
    \label{fig:cc_jepa_paper_trajectories}
\end{figure*}

\begin{table}
    \setlength{\tabcolsep}{2.788pt}
    \caption{Channel charting metrics for different encoders.}
    \centering
    \begin{tabular}{cccccc}
    \toprule
    & Metric & $\mathrm{CT} \rbrk{\uparrow}$ & $\mathrm{TW} \rbrk{\uparrow}$ & $\mathrm{KS} \rbrk{\downarrow}$ & $\mathrm{RD} \rbrk{\downarrow}$ \\
    \midrule
    \multirow{3}{*}{\rotatebox[origin=c]{90}{\shortstack{Encoder}}} 
        & No pre-training & $0.9820$ & $0.9845$ & $0.2604$ & $0.8925$ \\
        & Pre-trained with $d_{\text{ADP}}$ & $0.9866$ & $0.9882$ & $0.1398$ & $0.8252$ \\
        & Pre-trained with $d_{\text{G-fuse}}$ & $0.9942$ & $0.9939$ & $0.0815$ & $0.7251$ \\
    \bottomrule
    \end{tabular}
    \label{tab:cc_jepa_paper_cc_metrics}
    \vspace*{-0.2cm}
\end{table}

\subsection{Evaluation Setup}
To validate our proposed approach, we used the measured \gls{csi} data from the DICHASUS dataset~\cite{dataset-dichasus-cf0x}. 
The dataset is a collection of indoor channel measurements between a moving robot transmitter and four receiving arrays with eight antennas each. 
The system operates over $1024$ \gls{ofdm} subcarriers, spread over a $50$ MHz bandwith centered at a carrier frequency of $1.272$ GHz, and time is slotted to $40$ ms intervals.
We chose the three subsets \textit{dichasus-cf02}, \textit{dichasus-cf03} and \textit{dichasus-cf07} to form our training dataset, comprising $\datasetsize=80,000$ samples.
We left out the remaining samples for testing.

Our encoder is a \gls{mlp} with $5$ hidden layers of $1024$, $512$, $256$, $128$ and $64$ neurons, each followed by a ReLU activation.
We pre-process the channels similarly to~\cite{stephan2024angle}, which we skip detailing for brevity.
Unless stated otherwise, the predictor is a \gls{gru}~\cite{cho-etal-2014-learning} with a hidden layer of size $256$, and its recurrent state is fed to another two-layer \gls{mlp} to form its output.
The predictor transforms the sequence of input velocities, to a sequence of pseudo-velocities, which is sequentially added to the encoder's output to form the inferred target sequence.
The network is trained end-to-end with a learning rate of $0.005$, a batch size of $200$, and a \gls{ema} decay $\emadecay=0.99$.
We use weight decay with a fixed penalty factor of $0.0003$, and decrease the learning rate after every training epoch by a factor of $0.97$.
We train the network with a prediction horizon of $\predictionhorizon=300$ slots.


To test the quality of the learned latent space, we use four metrics from the channel charting literature: continuity $\rbrk{\mathrm{CT}}$, trustworthiness $\rbrk{\mathrm{TW}}$, Kruskal's stress $\rbrk{\mathrm{KS}}$ and Rajski's distance $\rbrk{\mathrm{RD}}$.
Due to the lack of space, we refer the reader to~\cite{studer2018channel}, \cite{stephan2024angle} for their detailed definitions.
All metrics take values between $0$ and $1$.
Both $\mathrm{CT}$ and $\mathrm{TW}$ are optimal at $1$, while $\mathrm{KS}$ and $\mathrm{RD}$ are optimal at $0$.

To evaluate the performance of the predictor, we divide the physical user locations in $10$ regions, and train a \gls{1nn} model to fit the region classes from a small subset of the learned chart.
Then, we use the predictor's output from an initial channel estimation and a velocity sequence to verify whether it follows the user's dynamics in the chart.

To examine the impact of pre-training, we use the encoder as the base of a siamese network, that learns a given channel distance, in the first stage.
We use two pre-training approaches, on dissimilarities proposed by~\cite{stephan2024angle}.
In the first one, the encoder is pre-trained on the \gls{adp} distance $d_{\text{ADP}}$~\cite[eq. (6)]{stephan2024angle} which does not yield a globally robust chart.
In the second one, we use the geodesic distance $d_{\text{G-fuse}}$~\cite[eq. (15)]{stephan2024angle} which produces a chart that preserve the original spatial geometry.

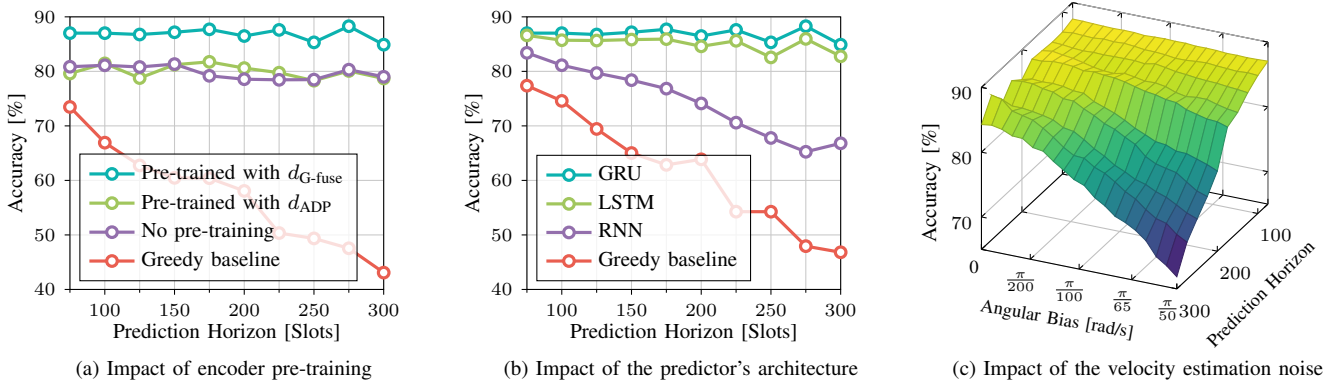
\begin{figure*}
    \centering
    \subfloat[Impact of encoder pre-training]{\parbox[b][4.6cm][t]{0.33\textwidth}{
        \begin{tikzpicture}[baseline,remember picture]

\definecolor{black}{RGB}{0,0,0}
\definecolor{grid}{RGB}{200,200,200}
\definecolor{pretrainedadp}{RGB}{161,198,93}
\definecolor{pretrainedfused}{RGB}{12,178,175}
\definecolor{random}{RGB}{147,111,172}
\definecolor{baselined}{RGB}{233,94,80}

\begin{axis}[
width=0.23\textwidth,
height=0.2\textwidth,
scale only axis,
legend cell align={left},
legend style={
  fill opacity=0.8,
  draw opacity=1,
  text opacity=1,
  at={(0.03,0.03)},
  anchor=south west,
  draw=black,
  font=\footnotesize
},
tick label style={font=\scriptsize},
major tick length=2pt,
minor tick length=1.25pt,
every tick/.style={thin},
tick align=outside,
tick pos=left,
x grid style={grid},
xlabel={\footnotesize{Prediction Horizon {[Slots]}}},
xmajorgrids, xminorgrids,
xmin=75, xmax=300,
xtick style={color=black},
xtick={100,150,200,250,300},
minor xtick={75,125,175,225,275},
y grid style={grid},
ylabel={\footnotesize{Accuracy {[\%]}}},
ymajorgrids,
ymin=40, ymax=90,
ytick style={color=black},
ytick={40,50,60,70,80,90},
xlabel shift = -4 pt,
ylabel shift = -4 pt,
]
\addplot [very thick, pretrainedfused, mark=*, mark size=2, mark options={solid,fill=white,fill opacity=1}]
table {%
75 87.0145456291105
100 87.0097450914502
125 86.7601171331189
150 87.1729633718977
175 87.7106235898421
200 86.4816859488263
225 87.6050117613173
250 85.3199558350535
275 88.281887571408
300 84.8975085209543
};
\addlegendentry{Pre-trained with $d_{\text{G-fuse}}$}
\addplot [very thick, pretrainedadp, mark=*, mark size=2, mark options={solid,fill=white,fill opacity=1}]
table {%
75 79.6073160193942
100 81.4603235562383
125 78.7720224665162
150 81.2154961355672
175 81.7483558158514
200 80.6010273150593
225 79.7561326868609
250 78.2919687004945
275 80.0729681724353
300 78.6904133262925
};
\addlegendentry{Pre-trained with $d_{\text{ADP}}$}
\addplot [very thick, random, mark=*, mark size=2, mark options={solid,fill=white,fill opacity=1}]
table {%
75 80.8506552733906
100 81.1146848447026
125 80.8026498967884
150 81.3355095770726
175 79.184868705295
200 78.5607988094667
225 78.4503864432816
250 78.4839902069032
275 80.308194517786
300 79.0024482742067
};
\addlegendentry{No pre-training}
\addplot [very thick, baselined, mark=*, mark size=2, mark options={solid,fill=white,fill opacity=1}]
table {%
75 73.4729241877256
100 66.9230769230769
125 62.7469879518072
150 60.4685990338164
175 60.4081632653061
200 58.0673076923077
225 50.3478260869565
250 49.3590361445783
275 47.5490909090909
300 43.0869565217391
};
\addlegendentry{Greedy baseline}
\end{axis}

\end{tikzpicture}
        \label{fig:cc_jepa_paper_downstream_pretrain}}}
    \hfill
    \subfloat[Impact of the predictor's architecture]{\parbox[b][4.6cm][t]{0.33\textwidth}{
        \begin{tikzpicture}[baseline,remember picture]

\definecolor{black}{RGB}{0,0,0}
\definecolor{grid}{RGB}{200,200,200}
\definecolor{pretrainedadp}{RGB}{161,198,93}
\definecolor{pretrainedfused}{RGB}{12,178,175}
\definecolor{random}{RGB}{147,111,172}
\definecolor{baselined}{RGB}{233,94,80}

\begin{axis}[
width=0.23\textwidth,
height=0.2\textwidth,
scale only axis,
legend cell align={left},
legend style={
  fill opacity=0.8,
  draw opacity=1,
  text opacity=1,
  at={(0.03,0.03)},
  anchor=south west,
  draw=black,
  font=\footnotesize
},
tick label style={font=\scriptsize},
major tick length=2pt,
minor tick length=1.25pt,
every tick/.style={thin},
tick align=outside,
tick pos=left,
x grid style={grid},
xlabel={\footnotesize{Prediction Horizon {[Slots]}}},
xmajorgrids, xminorgrids,
xmin=75, xmax=300,
xtick style={color=black},
xtick={100,150,200,250,300},
minor xtick={75,125,175,225,275},
y grid style={grid},
ylabel={\footnotesize{Accuracy {[\%]}}},
ymajorgrids,
ymin=40, ymax=90,
ytick style={color=black},
ytick={40,50,60,70,80,90},
xlabel shift = -4 pt,
ylabel shift = -4 pt,
]
\addplot [very thick, pretrainedfused, mark=*, mark size=2, mark options={solid,fill=white,fill opacity=1}]
table {%
75 87.0145456291105
100 87.0097450914502
125 86.7601171331189
150 87.1729633718977
175 87.7106235898421
200 86.4816859488263
225 87.6050117613173
250 85.3199558350535
275 88.281887571408
300 84.8975085209543
};
\addlegendentry{GRU}
\addplot [very thick, pretrainedadp, mark=*, mark size=2, mark options={solid,fill=white,fill opacity=1}]
table {%
75 86.5968988526715
100 85.7135999231914
125 85.65599347126879
150 85.82881282703663
175 85.91522250492055
200 84.5902741107004
225 85.6079880946666
250 82.55964668042821
275 85.95842734386251
300 82.76126926215737
};
\addlegendentry{LSTM}
\addplot [very thick, random, mark=*, mark size=2, mark options={solid,fill=white,fill opacity=1}]
table {%
75 83.39974077096635
100 81.12428592002304 
125 79.68892515961787 
150 78.38797945369882 
175 76.8326052517882 
200 74.11550093610484 
225 70.59190629350487 
250 67.78359176227737 
275 65.2489078776823 
300 66.80428207959291
};
\addlegendentry{RNN}
\addplot [very thick, baselined, mark=*, mark size=2, mark options={solid,fill=white,fill opacity=1}]
table {%
75 77.37665463297232
100 74.58173076923077 
125 69.44096385542169 
150 65.01932367149759 
175 62.84753901560625 
200 63.86538461538461 
225 54.26570048309178 
250 54.26024096385542 
275 47.94181818181818 
300 46.81642512077295 
};
\addlegendentry{Greedy baseline}
\end{axis}

\end{tikzpicture}
        \label{fig:cc_jepa_paper_downstream_predictor}}}
    \hfill
    \subfloat[Impact of the velocity estimation noise]{\parbox[b][4.6cm][t]{0.33\textwidth}{
        \begin{tikzpicture}[baseline,remember picture]
\begin{axis}[
width=0.21\textwidth,
height=0.21\textwidth,
scale only axis,
tick label style={font=\scriptsize},
major tick length=2pt,
every tick/.style={thin},
ylabel={\scriptsize{Prediction Horizon}},
xlabel={\scriptsize{Angular Bias [rad/s]}},
zlabel={\footnotesize{Accuracy [\%]}},
ylabel style={rotate=45},
xlabel style={rotate=-10},
y dir=reverse,
scaled x ticks = false,
xtick={0, 3.1416/200, 3.1416/100, 3.1416/65, 3.1416/50},
xticklabels={0,$\frac{\pi}{200}$,$\frac{\pi}{100}$,$\frac{\pi}{65}$,$\frac{\pi}{50}$},
zmin=65, zmax=90,
ztick={70,80,90},
ymajorgrids,xmajorgrids,zmajorgrids,
xlabel shift = -6 pt,
ylabel shift = -4 pt,
zlabel shift = -4 pt,
colormap name=viridis,
]

\addplot3 [surf] table {

    0.0000000000 75 87.4705967068
    0.0033069396 75 87.4513945562
    0.0066138793 75 87.3841870289
    0.0099208189 75 87.3265805770
    0.0132277585 75 87.2881762757
    0.0165346982 75 87.3313811147
    0.0198416378 75 87.3073784264
    0.0231485774 75 87.1681628342
    0.0264555171 75 87.2497719745
    0.0297624567 75 87.2977773511
    0.0330693964 75 87.2401708991
    0.0363763360 75 87.2641735874
    0.0396832756 75 87.1297585330
    0.0429902153 75 87.1489606836
    0.0462971549 75 87.1057558447
    0.0496040945 75 86.9809418655
    0.0529110342 75 87.0049445538
    0.0562179738 75 86.9521386395
    0.0595249134 75 86.9953434785
    0.0628318531 75 87.0625510057

    0.0000000000 100 86.8897316499
    0.0033069396 100 87.1009553070
    0.0066138793 100 87.0097450915
    0.0099208189 100 86.9425375642
    0.0132277585 100 87.1201574576
    0.0165346982 100 87.0001440161
    0.0198416378 100 87.0913542317
    0.0231485774 100 86.9473381019
    0.0264555171 100 86.7409149825
    0.0297624567 100 86.7169122942
    0.0330693964 100 86.6016993903
    0.0363763360 100 86.3760741203
    0.0396832756 100 86.4000768086
    0.0429902153 100 86.5392924007
    0.0462971549 100 86.6449042293
    0.0496040945 100 86.5008880995
    0.0529110342 100 86.4000768086
    0.0562179738 100 86.2848639048
    0.0595249134 100 86.4144784216
    0.0628318531 100 86.2944649801

    0.0000000000 125 86.7937208967
    0.0033069396 125 86.7169122942
    0.0066138793 125 86.5248907878
    0.0099208189 125 86.4144784216
    0.0132277585 125 86.5008880995
    0.0165346982 125 86.4432816475
    0.0198416378 125 86.4912870241
    0.0231485774 125 86.5104891748
    0.0264555171 125 86.5152897124
    0.0297624567 125 86.3376698190
    0.0330693964 125 86.6689069176
    0.0363763360 125 86.4096778839
    0.0396832756 125 86.1072440113
    0.0429902153 125 85.9728289568
    0.0462971549 125 85.7808074504
    0.0496040945 125 85.6655945466
    0.0529110342 125 85.5887859440
    0.0562179738 125 85.4111660506
    0.0595249134 125 85.2815515338
    0.0628318531 125 85.0943305650

    0.0000000000 150 86.9473381019
    0.0033069396 150 87.1633622966
    0.0066138793 150 86.9089338006
    0.0099208189 150 86.7121117565
    0.0132277585 150 86.3616725073
    0.0165346982 150 85.9776294945
    0.0198416378 150 85.7472036868
    0.0231485774 150 85.6511929336
    0.0264555171 150 86.0064327205
    0.0297624567 150 85.7472036868
    0.0330693964 150 85.4447698142
    0.0363763360 150 85.5023762661
    0.0396832756 150 85.2383466948
    0.0429902153 150 84.6574816379
    0.0462971549 150 84.3454466900
    0.0496040945 150 84.5470692718
    0.0529110342 150 84.5614708847
    0.0562179738 150 84.0958187317
    0.0595249134 150 83.7309778695
    0.0628318531 150 83.4957515242

    0.0000000000 175 87.0865536940
    0.0033069396 175 87.0769526187
    0.0066138793 175 87.3649848783
    0.0099208189 175 87.3697854160
    0.0132277585 175 87.2353703615
    0.0165346982 175 86.8849311123
    0.0198416378 175 87.0481493927
    0.0231485774 175 86.8513273487
    0.0264555171 175 86.7601171331
    0.0297624567 175 86.4624837982
    0.0330693964 175 86.0400364841
    0.0363763360 175 85.6463923959
    0.0396832756 175 85.4351687389
    0.0429902153 175 85.0223225001
    0.0462971549 175 84.8735058327
    0.0496040945 175 84.1102203447
    0.0529110342 175 83.7021746436
    0.0562179738 175 83.3085305554
    0.0595249134 175 83.2029187269
    0.0628318531 175 82.6988622726

    0.0000000000 200 85.3775622870
    0.0033069396 200 85.7904085258
    0.0066138793 200 85.6655945466
    0.0099208189 200 85.7472036868
    0.0132277585 200 85.6127886323
    0.0165346982 200 85.2575488455
    0.0198416378 200 85.1951418559
    0.0231485774 200 84.2302337862
    0.0264555171 200 83.6685708799
    0.0297624567 200 83.0829052854
    0.0330693964 200 82.3052181844
    0.0363763360 200 81.3979165667
    0.0396832756 200 80.7114396812
    0.0429902153 200 79.9721568816
    0.0462971549 200 79.7897364505
    0.0496040945 200 79.1512649417
    0.0529110342 200 79.0936584898
    0.0562179738 200 78.5752004224
    0.0595249134 200 78.0423407422
    0.0628318531 200 77.3702654697

    0.0000000000 225 86.6785079929
    0.0033069396 225 86.8801305746
    0.0066138793 225 86.8417262733
    0.0099208189 225 86.4096778839
    0.0132277585 225 86.1840526139
    0.0165346982 225 85.4927751908
    0.0198416378 225 85.4447698142
    0.0231485774 225 84.5662714224
    0.0264555171 225 83.8461907734
    0.0297624567 225 82.9388891556
    0.0330693964 225 82.3724257117
    0.0363763360 225 81.4747251692
    0.0396832756 225 80.6826364553
    0.0429902153 225 79.1368633287
    0.0462971549 225 78.3207719265
    0.0496040945 225 77.9367289136
    0.0529110342 225 77.6006912774
    0.0562179738 225 76.4821660026
    0.0595249134 225 75.4164466420
    0.0628318531 225 74.8739858864

    0.0000000000 250 84.4846622822
    0.0033069396 250 84.5422687341
    0.0066138793 250 84.5278671211
    0.0099208189 250 84.3214440017
    0.0132277585 250 84.0382122798
    0.0165346982 250 83.6685708799
    0.0198416378 250 82.7900724881
    0.0231485774 250 82.5260429168
    0.0264555171 250 81.5851375354
    0.0297624567 250 80.4954154865
    0.0330693964 250 79.6553213960
    0.0363763360 250 78.5800009601
    0.0396832756 250 77.8887235370
    0.0429902153 250 77.0294272959
    0.0462971549 250 76.5877778311
    0.0496040945 250 75.7956891172
    0.0529110342 250 75.2628294369
    0.0562179738 250 74.3987326581
    0.0595249134 250 73.4818299650
    0.0628318531 250 72.8769622198

    0.0000000000 275 87.3457827277
    0.0033069396 275 87.0145456291
    0.0066138793 275 85.5359800298
    0.0099208189 275 85.1567375546
    0.0132277585 275 84.7726945418
    0.0165346982 275 83.8461907734
    0.0198416378 275 82.7708703375
    0.0231485774 275 81.5275310835
    0.0264555171 275 80.5914262397
    0.0297624567 275 80.4906149489
    0.0330693964 275 79.4248955883
    0.0363763360 275 78.8056262301
    0.0396832756 275 77.7879122462
    0.0429902153 275 76.6213815947
    0.0462971549 275 75.7956891172
    0.0496040945 275 74.4851423359
    0.0529110342 275 73.4194229754
    0.0562179738 275 72.9921751236
    0.0595249134 275 71.6096202775
    0.0628318531 275 70.7455234986

    0.0000000000 300 84.2638375498
    0.0033069396 300 84.4702606692
    0.0066138793 300 84.1534251836
    0.0099208189 300 83.4669482982
    0.0132277585 300 83.3853391580
    0.0165346982 300 82.3148192598
    0.0198416378 300 82.0699918391
    0.0231485774 300 80.8266525851
    0.0264555171 300 79.7705342998
    0.0297624567 300 78.6664106380
    0.0330693964 300 78.0567423551
    0.0363763360 300 76.5349719169
    0.0396832756 300 75.5652633095
    0.0429902153 300 74.2643176036
    0.0462971549 300 73.6162450194
    0.0496040945 300 72.7281455523
    0.0529110342 300 71.4656041477
    0.0562179738 300 69.3293648889
    0.0595249134 300 68.2540444530
    0.0628318531 300 66.8666890692

};
\end{axis}
\end{tikzpicture}
        \label{fig:cc_jepa_paper_downstream_noise}}}
    \caption{Downstream task numerical results.}
    \label{fig:cc_jepa_paper_downstream_task}
\end{figure*}

\subsection{Discussion}
We present the learned channel representations in Fig.~\ref{fig:cc_jepa_paper_charts}.
In Fig.~\ref{fig:cc_jepa_paper_gt_locations}, we show the ground truth user positions, while Figs.~\ref{fig:cc_jepa_paper_random_chart}, \ref{fig:cc_jepa_paper_local_chart} and \ref{fig:cc_jepa_paper_global_chart} display the channel latent space learned by the proposed \gls{wjepa} without encoder pre-training, and with an encoder pre-trained with $d_{\text{ADP}}$ and $d_{\text{G-fuse}}$ respectively.
Gradient coloring is used to distinguish local neighborhoods.
Remarkably, all charts preserve the overall spatial geometry, with an increasing quality depending on the pre-training.
However, even the randomly initialized encoder and the one pre-trained with $d_{\text{ADP}}$ produce globally robust charts.
This is due to the idea behind our \gls{wjepa}: we task the encoder with learning channel embeddings that are predicable from each other given the velocity.
Since the model learn the user's dynamics in the environment, the obtained charts maintain the original user locations' structure.
We show the latent space quality metrics in Table~\ref{tab:cc_jepa_paper_cc_metrics}, which confirm the visual presentation of Fig.~\ref{fig:cc_jepa_paper_charts}.

Fig.~\ref{fig:cc_jepa_paper_gt_traj} shows a sequence of $30$ user trajectories, gradient-colored from beginning to end, each starting with a yellow cross indicating a channel estimation.
This channel is fed to the encoder, while the predictor takes the obtained chart point with the sequence of the user's velocity to autoregressively predict the next chart points.
Figs.~\ref{fig:cc_jepa_paper_random_traj}, \ref{fig:cc_jepa_paper_local_traj}, and \ref{fig:cc_jepa_paper_global_traj} depict the dynamics learned by the predictor, for the aforementioned pre-training approaches respectively.
We also show the corresponding encoder's channel embeddings, i.e. what the predictor should infer, assuming the channel is estimated.
Here, we notice an increase in the learned dynamics quality when the encoder is pre-trained with a globally robust channel distance.
This is explained by the fact that the encoder pre-trained with $d_{\text{G-fuse}}$ facilitates the predictor's task, as it is able to produce channel embeddings that simulate the user's location.
Hence the corresponding \gls{jepa} network produces the best, and most visually pleasing, trajectory predictions.
This highlights a trade-off between channel charting pre-training and learned latent space quality.

Fig.~\ref{fig:cc_jepa_paper_downstream_pretrain} plots the downstream task accuracy for the proposed \glspl{wjepa} with and without encoder pre-training, for different prediction horizons.
We compare our approach with a greedy baseline that labels the user to the same region for the entire trajectory length.
We notice that all our proposed methods maintain an almost constant accuracy of $87\%$ when the encoder is pre-trained with $d_{\text{G-fuse}}$ and $80\%$ when $d_{\text{ADP}}$ is used in pre-training or the network is trained from scratch.
The baseline's accuracy falls from $73\%$ to $43\%$ when the prediction horizon goes from $75$ to $300$ slots.
Hence, our methods display a two-fold increase in accuracy for longer look ahead predictions.

Fig.~\ref{fig:cc_jepa_paper_downstream_predictor} presents the downstream task accuracy for different predictor networks.
We compare standard \gls{rnn}, \gls{gru}, and \gls{lstm}~\cite{sepp1997lstm} modules.
We observe that while the \gls{gru} and the \gls{lstm} demonstrate constant accuracy while modelling long time series, the \gls{rnn} predictor's accuracy decreases from $80\%$ to less than $70\%$ when the prediction horizon increases from $100$ to $300$ slots.

Finally, we test our \gls{wjepa}'s prediction accuracy on the downstream task in the presence of velocity estimation noise, as shown in Fig~\ref{fig:cc_jepa_paper_downstream_noise}.
We simulate this estimation noise as a bias in the angular velocity.
We notice that for short prediction horizons, below $150$ slots, the performance is almost unaffected by noise, with an accuracy always above $85\%$.
However, for long prediction sequence such as $300$ slots, the performance starts quickly degrading, decreasing from $85\%$ without any noise to less than $70\%$ for when the angular bias exceeds $\frac{\pi}{65}$ rad/s.

\section{Conclusion}\label{section:ccjepa_6}
In this paper, we presented a novel \gls{ml} technique to learn the dynamics of \gls{csi} data in a self-supervised manner.
We proposed a \gls{jepa} whose encoder mimics a channel charting function, mapping \gls{csi} data to pseudo-locations, while the predictor is conditioned on the user's velocity to output future channel embeddings.
We demonstrated extensive results, showcasing different trade-offs of the proposed architecture.
Building on the promising performance of \gls{jepa} on wireless modalities, our future works will explore its benefits for resource allocation and scheduling problems.

\bibliography{references}
\bibliographystyle{IEEEtran}

\end{document}